\documentclass[sn-mathphys-num,iicol]{sn-jnl}

\usepackage{tikz}
\usepackage{color}
\definecolor{darkgreen}{RGB}{0,127,0}
\definecolor{darkblue}{RGB}{0,0,175}

\newcommand{\addedtext}[2]{{{#2}}}
\newcommand{\modifiedtext}[2]{{{#2}}}
\newcommand{\deletedtext}[2]{{}}

\usepackage{graphicx}%
\usepackage{multirow}%
\usepackage{amsmath,amssymb,amsfonts}%
\usepackage{amsthm}%
\usepackage{mathrsfs}%
\usepackage[title]{appendix}%
\usepackage{xcolor}%
\usepackage{textcomp}%
\usepackage{manyfoot}%
\usepackage{booktabs}%
\usepackage{algorithm}%
\usepackage{algorithmicx}%
\usepackage{algpseudocode}%
\usepackage{listings}%

\usepackage{multicol}
\usepackage{balance}
\usepackage{graphics}
\usepackage{amsmath}

\usepackage{amssymb}
\usepackage{epsfig} 
\usepackage{booktabs}
\usepackage{float}
\usepackage{url}
\usepackage{gensymb}
\usepackage{hyperref}
\usepackage{makecell}
\usepackage[colorinlistoftodos]{todonotes}
\usepackage{wrapfig}
\usepackage{array,multirow,graphicx}
\usepackage{xspace}
\def\OURS{GEOTACT\xspace}
\theoremstyle{thmstyleone}%
\theoremstyle{thmstyletwo}%

\theoremstyle{thmstylethree}%

\begin{document}

\title[Tactile-based Object Retrieval From Granular Media]{Tactile-based Object Retrieval From Granular Media}


\author*[1]{\fnm{Jingxi} \sur{Xu}}\email{jxu@cs.columbia.edu}
\equalcont{These authors contributed equally to this work.}

\author[2]{\fnm{Yinsen} \sur{Jia}}
\equalcont{These authors contributed equally to this work.}

\author[3]{\fnm{Dongxiao} \sur{Yang}}
\equalcont{These authors contributed equally to this work.}

\author[1]{\fnm{Patrick} \sur{Meng}}

\author[1]{\fnm{Xinyue} \sur{Zhu}}

\author[3]{\fnm{Zihan} \sur{Guo}}

\author[1,4]{\fnm{Shuran} \sur{Song}}

\author[3]{\fnm{Matei} \sur{Ciocarlie}}

\affil[1]{\orgdiv{Department of Computer Science}, \orgname{Columbia University}, \orgaddress{\city{New York}, \postcode{10027}, \state{NY}, \country{USA}}}

\affil[2]{\orgdiv{Department of Electrical Engineering}, \orgname{Columbia University}, \orgaddress{\city{New York}, \postcode{10027}, \state{NY}, \country{USA}}}

\affil[3]{\orgdiv{Department of Mechanical Engineering}, \orgname{Columbia University}, \orgaddress{\city{New York}, \postcode{10027}, \state{NY}, \country{USA}}}

\affil[4]{\orgdiv{Department of Electrical Engineering}, \orgname{Stanford University}, \orgaddress{\city{Stanford}, \postcode{94305}, \state{CA}, \country{USA}}}


\abstract{We introduce \OURS, the first robotic system capable of grasping and retrieving objects of potentially unknown shapes buried in a granular environment. While important in many applications, ranging from mining and exploration to search and rescue, this type of interaction with granular media is difficult due to the uncertainty stemming from visual occlusion and noisy contact signals. To address these challenges, we use a learning method relying exclusively on touch feedback, trained end-to-end with simulated sensor noise. We show that our problem formulation leads to the natural emergence of learned pushing behaviors that the manipulator uses to reduce uncertainty and funnel the object to a stable grasp despite spurious and noisy tactile readings. We introduce a training curriculum that bootstraps learning in simulated granular environments, enabling zero-shot transfer to real hardware. Despite being trained only on seven objects with primitive shapes, our method is shown to successfully retrieve 35 different objects, including rigid, deformable, and articulated objects with complex shapes.}




\maketitle

\section{Introduction}

Interacting with granular media, for example, in order to retrieve buried objects, is a foundational skill for numerous important robotics applications. Examples include mining and demining, search and rescue, excavation and construction, archaeology, and exploration of terrestrial, seabed, and even extraterrestrial environments, all requiring robots that can extract valuable minerals, metals, or other geological materials from the ground.  

Many of these aforementioned applications require operation in hazardous environments that pose a significant danger to human workers. For example, mining or rescue sites are often located in remote and inhospitable areas. In the United States, about 200 casualties occur per year from cave-ins, ground collapses, or other excavation incidents alone~\citep{lew2002excavation}. Looking further out, future extraterrestrial sites that are currently being contemplated, such as the moon~\citep{creech2022artemis} or even Mars~\citep{zhang2022analysis,nasamars}, could completely prevent on-site operators. As a result, enabling autonomous robots to replace humans in searching through loose soil or other granular media could play an important role in reducing exposure to injuries. However, while humans are experts at performing such tasks, for example, using tactile sensing to compensate when visual feedback is absent, transferring such behaviors to robots poses multiple challenges. 

First, uncertainty and sensing noise greatly affect operation inside sensor-deprived environments such as granular media. With vision unavailable, touch becomes the prominent sensing modality. However, all tactile sensors exhibit some level of noise in their readings, which is typically exacerbated when submerged under granular media. Tactile sensing is fundamentally an active sensing modality, needing an intelligent exploration policy to guide movement and compensate for the sparse nature of the data. 


Second, manipulation within granular media places specific requirements on the hardware. In a granular environment, contacts come from every angle and direction, thus requiring extensive sensing coverage. Despite many breakthroughs~\citep{patel2021digger,lambeta2020digit,johnson2009retrographic,yuan2017gelsight,dong2017improved,chang2023investigation} in sensor developments, many sensors are either flat or do not have sensing coverage around edges and corners.  
In addition, we want an intruder shape that can render low resistance and smooth movement inside granular media. For example, circular and elliptical shapes typically experience slightly less resistance during horizontal dragging and significantly less resistance during vertical lifting compared to rectangular shapes~\cite{tripura2022role}.
Due to such compounding limitations, robotic manipulation in granular media remains poorly explored compared to tactile-based work with objects in open-air environments.

\begin{figure}[!t]
\centering
\includegraphics[width=\linewidth]{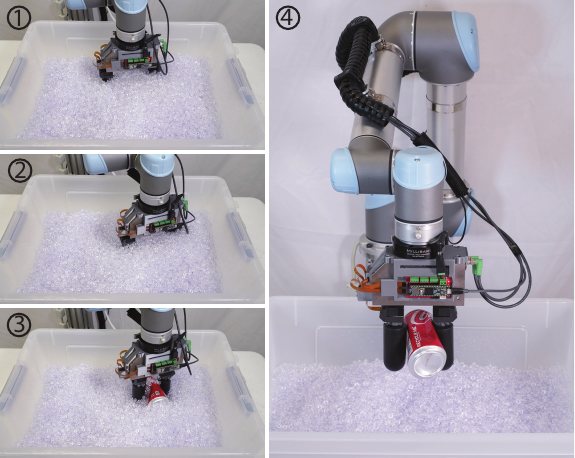}
\caption{\textbf{Retrieving buried object from granular media.} In this task, an object is buried under granular beads. We mount a parallel gripper with two tactile fingers on a robot arm to retrieve the buried object using only tactile feedback. Our learned policy uses a series of pushing actions to funnel the buried object into a stable grasp and then successfully retrieve the target object.}
\label{fig:teaser}
\end{figure}

In this work, we propose \textbf{\OURS} (\textbf{G}ranular \textbf{E}nvironment \textbf{O}bject \textbf{T}racing and \textbf{AC}quisition via \textbf{T}ouch), a novel method for retrieving objects buried under granular media. \OURS relies exclusively on tactile sensing, as we assume the object can be completely hidden from any vision sensors. In particular, we use the DISCO tactile finger~\citep{piacenza2020sensorized}, which provides both the streamlined shape and sensing coverage needed to enable the task. 

Still, sensor noise remains a significant challenge in addressing our proposed task. To address this, we take a multi-pronged approach. \OURS is learning-based, trained with reinforcement learning entirely in simulation, and then zero-shot transferred onto the real hardware. Our method is trained end-to-end, without separating perception and control policies, and is model-free, making no assumptions about the shapes or dynamics of objects. Compared to heuristics-based methods, which are more sensitive to sensor noise, we find that learning-based methods can incorporate noise into the training process and thus become more robust to the considerable amount of noise in granular media.

Critically, \OURS relies on a pushing behavior which further reduces uncertainty and increases robustness against noisy tactile readings. Despite pushing having been shown in the literature to be a key element of many manipulator operations~\citep{brost1988automatic,mason1986mechanics,dogar2010push,dogar2012planning}, no previous work has utilized pushing action to grasp objects within granular media. The \OURS formulation of the learning problem action space leads to the emergence of a series of pushing behaviors that funnels the target object into a stable grasp. 

To enable this learning process, we propose a curriculum strategy that enables learning in simulation, followed by zero-shot transfer to real hardware. We first pre-train our policy in an open-air environment and then fine-tune it inside simulated granular media. Despite the recent effort from the community to improve granular material simulation accuracy and efficiency~\citep{kloss2012models,millard2023granular,thompson2022lammps,xian2023fluidlab,fang2021chrono,servin2021multiscale,ansys,haeri2020efficient,hu2019difftaichi}, granular media simulation are still significantly slower than open-air simulation. However, our curriculum strategy helps to bootstrap this training process, allowing us to reach a better policy within a shorter training time by transferring the skills learned from the open air to granular media. 

In summary, our main contributions are as follows:
\begin{itemize}
    \item 
    To the best of our knowledge, we are the first to demonstrate robotic grasping of objects buried under a specific type of granular media using only touch sensing.
    \item In order to handle the tactile noise inherent to this problem, we use a learning-based method trained end-to-end in the presence of noise, and introduce a curriculum that enables training in simulation and zero-shot transfer to real hardware for this task.
    \item We show that our formulation leads to the emergence of pushing behaviors that reduce uncertainty and enable robust performance in real scenarios.
\end{itemize}

\modifiedtext{1-1}{Our quantitative real-robot experiments show that \OURS achieves success rates of 54\% and 63\%  on its training set (seven objects) and testing set (six additional objects not seen during training), respectively.} 
In additional testing, we also show promising early results on an extended set of 22 objects, including rigid, deformable, and articulated objects with various complex shapes. Our extensive experiments in both simulation and on the real robot show that our method is more robust to sensor noise and achieves a higher success rate compared to baselines. 
\section{Related Work}

\noindent \textbf{Robot Interaction with Granular Media}.  
Unlike manipulating objects buried inside granular media, there are many more works focusing on manipulating granular materials directly, such as scooping and
bulldozing~\cite{sarata2004trajectory,millard2023granular,zhu2023few}, plate dragging (soil-tool interaction)~\cite{kobayakawa2020interaction,swick1988model,gravish2014force}, and pouring~\cite{wu2020squirl,yamaguchi2015pouring,matl2020inferring,tuomainen2022manipulation}. In addition to tactile feedback, some works~\cite{clarke2018learning,clarke2019robot} also incorporate visual and audio feedback for scooping and pouring tasks. Other works~\cite{li2013terradynamics,hauser2016friction,zhu2019data} study legged locomotion on granular materials. In terms of hardware design, there are works that use granular materials and particle jamming to design gripper~\cite{brown2010universal,fakhri2023systematic}, manipulator~\cite{cheng2012design}, swarm robots~\cite{karimi2020boundary}, or haptic displays~\cite{brown2020soft,stanley2013haptic}.

In contrast, interacting with objects buried inside granular materials has only recently begun receiving attention
from the research community, but is \textit{only} limited to tactile perception such as object mapping, localization or identification. Jia, et al.~\cite{jia2021tactile} teleoperate a robotic hand with the BioTac tactile sensors to detect contact with a
cylinder fixed inside a bed of granular media. They take advantage of the multi-modal sensors to distinguish between contact with an object embedded in granular media and ubiquitous contact with surrounding granular media. In their follow-up work~\cite{jia2022autonomous}, instead of teleoperation, they develop a tactile exploration policy for efficient localization and mapping of the buried object. Patel, et al.~\cite{patel2021digger} develop a new tactile finger built on top of GelSight technology~\cite{johnson2009retrographic,johnson2011microgeometry} that facilitates penetration in granular media with the help of mechanical vibrations. They use this sensor to identify four simple 2D shapes i.e. triangle, square, hexagon, and circle. More recently, Zhang, et al.~\cite{zhang2023grains} propose a proximity sensing system to detect an object's presence without contact and finally use a Bayesian-optimization-algorithm-guided exploration strategy to localize buried objects. Different from all these aforementioned works, \OURS goes beyond tactile perception and focuses on retrieving any buried objects. To the best of our knowledge, ours is the first work to enable a robotic gripper to retrieve buried objects from granular media.

\vspace{3mm}
\noindent \textbf{Robotic Pushing for Grasping}.
Pushing is a key element of many manipulator operations. It is a good way to reduce uncertainty in the locations of objects, to move many objects at once, and to move objects that are hard to grasp. Combining both non-prehensile (e.g., pushing) and prehensile (e.g., grasping) manipulation policies has received increasing attention from the community. Brost~\cite{brost1988automatic} presents an algorithm that plans parallel-jaw push-grasping motions for polygonal objects with pose uncertainty, where the object is pushed by one plate towards the second one, and then squeezed between the two. Mason~\cite{mason1986mechanics} investigates the mechanics and planning of pushing in object manipulation under uncertainty. The seminal work of Dogar et al.~\cite{dogar2010push,dogar2012planning} presents a robust planning framework for push-grasping (non-prehensile motions baked
within grasping primitives) to reduce grasp uncertainty.

While the policies in the works above remain largely hand-crafted, other methods~\cite{omrvcen2009autonomous,clavera2017policy} explore the model-free planning of pushing motions to move objects to target positions that are more favorable for pre-designed grasping algorithms. Boularias et al.~\cite{boularias2015learning} explores the use of reinforcement learning for training control policies to select among push and grasp proposals represented by hand-crafted features. However, their method models perception and control policies separately (not end-to-end); it relies on model-based simulation to predict the motion of pushed objects and to infer its benefits for future grasping. Similarly, Yu et al.~\cite{yu2023novel} also separates push planning and grasp planning by training two networks. More recently, some works~\cite{zeng2018learning,yang2021collaborative,deng2019deep,tang2021learning} have been training model-free end-to-end policy using reinforcement learning to push and separate objects in clutter for better grasping. Our method is also model-free and trained end-to-end, making no assumptions about the shapes or dynamics of the objects. However, unlike all the previous works cited here, ours is the first to demonstrate a pushing behavior for grasping inside granular media.

\vspace{3mm}
\noindent \textbf{Active Tactile Exploration}.
Tactile sensors need an intelligent exploration policy to actively interact with the target object and efficiently gather sparse and local tactile readings. These exploration policies can be roughly divided into heuristics-based and learning-based. One of the most popular heuristics-based exploration policies is contour-following~\cite{martinez2013active,yu2015shape,suresh2021tactile}. Other heuristics include information gain (uncertainty reduction)~\cite{hebert2013next,xu2013tactile,schneider2009object,driess2017active}, attention cubes~\cite{rajeswar2021touch}, Monte Carlo tree search~\cite{zhang2017active}, and dynamic potential fields~\cite{bierbaum2009grasp}. Some works study tactile exploration policy specifically for decision making, such as object identification, where they pre-train a discriminator with a pre-collected dataset and then use it to estimate action quality with Bayesian methods to reduce uncertainty~\cite{fishel2012bayesian, lepora2013active, martinez2017active, kaboli2017tactile, kaboli2019tactile}. Depending on the task, heuristic-based exploration policies can improve efficiency and require no training, but their performance can deteriorate significantly when sensor noise shows up. 

In contrast, Xu et al.~\cite{xu2022tandem,xu2023tandem3d} study learning-based exploration policy trained through reinforcement learning in an unsupervised fashion. They also aim for object identification, and their method has two separate components, trained jointly: a discriminator for decision-making and an explorer for guiding the exploration. Different from these previous works, our method aims to grasp buried objects inside granular media, which is a more challenging and novel task. In addition, our method only has one end-to-end policy that guides both the pushing and grasping actions. 

\section{Method} 

We formulate our problem as follows. A robotic manipulator is presented with a workspace filled with granular media. 
An object is buried completely under the granular media, placed with unknown orientation or exact position, such that the object is in the way of the gripper and a contact is guaranteed if the gripper moves straight forward.
The goal of the manipulator is to retrieve the buried object, by lifting it clear of the granular environment. We assume that the object can start out fully buried with no part visible from the surface. Our goal is to perform the task without relying on vision and using only touch sensing.

Tactile sensing in granular media requires the ability to distinguish between readings due to ubiquitous contacts with the granules themselves, and readings indicating contact with the target object. As explained later in this section, we use a simple filtering technique to make this distinction. However, the nature of operation in such environments increases the noise inherent in any tactile sensor. We thus need an exploration policy that is well-suited for handling tactile noise and spurious or erroneous readings. 

Rather than attempting to handle sensor noise via heuristics or further signal processing, we achieve a high level of robustness by formulating the object retrieval task as a learning problem, trained in the presence of noise. We describe this formulation next, and show how it leads to the natural emergence of a pushing behavior that has the role of reducing uncertainty before a grasp is attempted. Finally, we will discuss our training curriculum, which enables training of this policy in simulation followed by zero-shot transfer to hardware.

\subsection{Action Space Definition and Pushing Behavior} 
\label{sec:action}

We formulate our problem as a model-free reinforcement learning task. 
Object exploration and retrieval are achieved through a series of discrete translation/rotation actions of the gripper, where the translation action is two-dimensional on a 2D $xy$ plane, and the rotation action is one-dimensional along the $z$ axis.
At each step, our policy takes in a sequence of contact locations and forces from the past steps and then outputs the next action for the gripper to take for exploration. Our policy also determines when to stop pushing and execute the grasp. 


In defining the action space of our learning task, we build on the extensive body of work (reviewed earlier) showing that a planar-pushing primitive can be used to reduce uncertainty and funnel an object into a state that lends itself to robust grasping. However, a heuristic-based pushing primitive is difficult to implement in very noisy conditions. Instead, we aim to formulate the learning problem such that the agent naturally learns to make use of pushing in the context of end-to-end task learning.

Specifically, our action primitives consist of a planar pushing action and a grasping action. The gripper moves in a 2D plane and is always perpendicular to the table, as shown in Fig.~\ref{fig:teaser}. 
Given the current gripper pose (position $[x, y]$ and rotation $\theta$), the action is a 4-dimensional vector $[dx, dy, d\theta, G]$.
A value of $G=0$ indicates a pushing action, in which case $dx, dy, d\theta$ indicate the change to $x$, $y$, and $\theta$, respectively. For $x, y$, the change is 1cm and for $\theta$, the change is $15^\circ$; namely, $dx, dy \in \{-1\text{cm}, 0, +1\text{cm}\}$ and $d\theta \in \{-15^\circ, 0, +15^\circ\}$. 
Notice that in this action space formulation, it is possible to change $x, y$, and $\theta$ simultaneously in one action, by setting the corresponding elements at the same time.
$G=1$ corresponds to a grasping action. In this case, the values of $dx, dy, d\theta$ are ignored, the gripper is closed, and the object is lifted. The episode is terminated when a grasping action is executed or a predefined maximum number of steps is reached. 

As discussed in the results section, this formulation results in the learned policy effectively making use of sequences of pushing behaviors of various lengths. However, a critical aspect of learning such behaviors is training in the presence of noise, which we discuss later in Sec.~\ref{sec:noise}.

\subsection{Observation Space and Tactile Data Processing} 
\label{sec:noise_gap}

\begin{figure}[t]
\centering
\includegraphics[width=\linewidth]{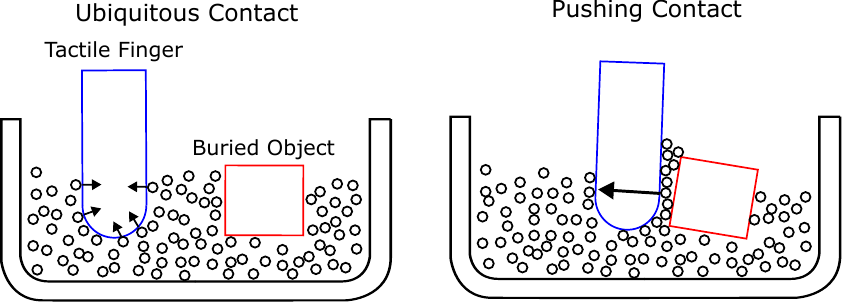}
\caption{\textbf{Two types of contact inside granular media: ubiquitous contact and pushing contact.} Ubiquitous contacts are between the finger and the granular media, but pushing contacts are sensed only when the finger has started pushing the buried object.}
\label{fig:pushing_contact}
\end{figure}

\begin{figure*}[t]
\centering
\includegraphics[width=\textwidth]{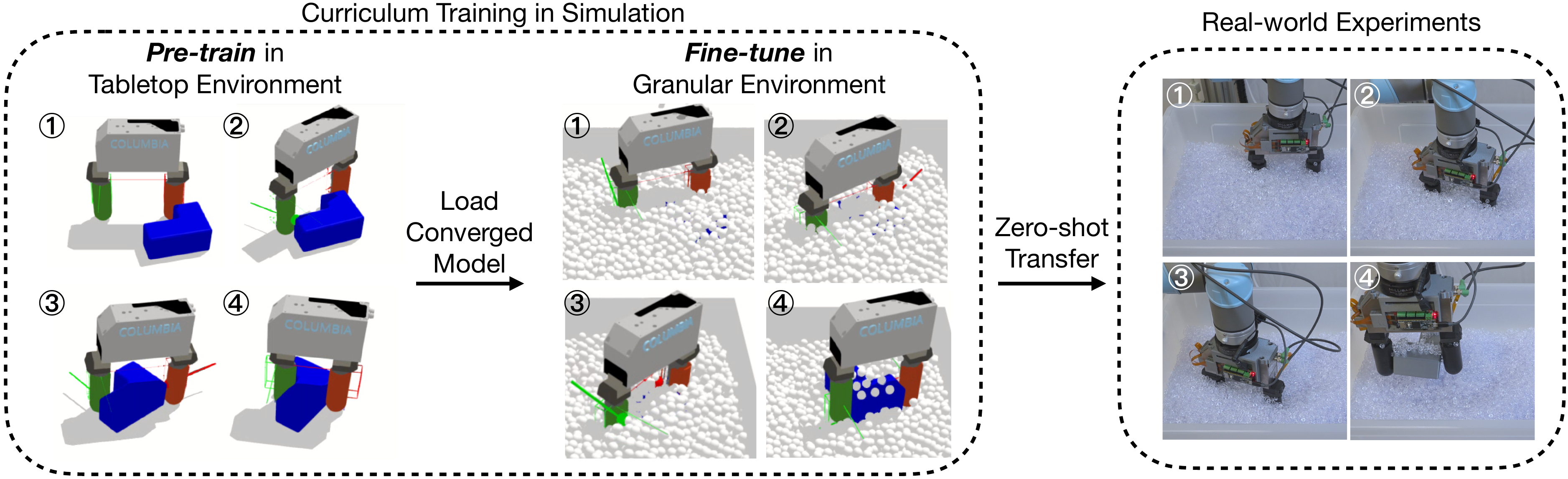}
\caption{\textbf{Curriculum training strategy.} 
The numbering of the frames corresponds to their order within a single episode.
We first pre-train our policy on a tabletop environment and then fine-tune it inside granular media. We then zero-shot deploy our policy on the real hardware for evaluation. This curriculum allows the policy to converge to a higher performance within a shorter training time.}
\label{fig:curriculum}
\end{figure*}

At each time step $t$, the observation of our policy is a sequence of contact locations $p_t, p_{t-1}, \dots, p_{t-h}$, net forces $f_t, f_{t-1}, \dots, f_{t-h}$, and previous actions $a_{t-1}, a_{t-2}, \dots, a_{t-h-1}$ from the last $h = 10$ steps. Observations from both fingers are concatenated to form the final observation of the policy. Both the contact location $p_t$ and the net force $f_t$ are 3-dimensional vectors under the finger's frame of reference (using the center of the finger base as the origin). The previous action $a_{t-1}$ is a 4-dimensional vector, as discussed in Sec.~\ref{sec:action}.
When there is no contact recorded after a particular step, we simply fill in zeros in the location and force vectors.
Compared to high-dimensional or multimodal tactile data, contact locations and forces are easy to simulate, enabling a large amount of training in simulation and zero-shot transfer to real-world experiments.

Unlike open-air environments, where the contacts are simply between the tactile finger and the target object, inside the granular media there are two types of contacts our policy needs to distinguish between: ubiquitous contact and pushing contact (Fig.~\ref{fig:pushing_contact}). Ubiquitous contact is the contact with the granular media, and the contact forces are small whether the finger is moving inside the granular media or not. Pushing contact is the contact sensed by the finger when the object has started moving due to the pushing, and because of the larger dimensions of the target object compared with granular media, the contact forces are much larger. We assume only pushing contacts carry important information for our policy, and we want to discard ubiquitous contacts with granular media. We thus use a force filter to keep only the contact positions with a force magnitude larger than a threshold. 
The net force vector is the combined force applied to our finger, which also contains the direction of the force.

Tactile observation in granular media is much more noisy compared to open air. First, there might not be direct contact between the finger and the target object when pushing contacts are detected. Instead, there could be a layer of granules between the finger and the target object. This layer can potentially introduce more noise to the force directions. Second, for small and light objects, pushing contact can be hard to distinguish from ubiquitous contacts, and, depending on how fast the finger accelerates and moves inside the medium, pushing contacts can be falsely reported even though the finger has not started pushing the target object. 

Ideally, object- and granular media-specific force filtering thresholds might be needed for better filtering accuracy. 
However, in this work, the filtering threshold only differs between simulation and real hardware but remains the same for all objects, and we observe that our policy learns to be robust to spurious pushing contacts.
The filtering threshold is found experimentally through calibration. During the calibration phase, we drag the submerged finger through the granular media without any buried objects. The dragging follows a predefined trajectory with multiple changes in direction. We record the maximum force sensed during the whole trajectory. We then find the minimum integer value larger than the maximum force to be the chosen force threshold. We do the calibration process separately in simulation and on the real hardware, giving us different filtering thresholds.

\subsection{Reward Design and Policy Architecture}

The reward to our policy is a hybrid of dense and sparse rewards. We constantly reward the policy when the center of the gripper gets closer to the center of the object, and we provide a large sparse reward when the policy decides to grasp and successfully lift the target object from the granular materials. At step $t$, let $d_t$ denote the Euclidean distance between the gripper center position $(x_g, y_g)$ and the object center position $(x_o, y_o)$, then the reward at each step $t$ is computed as 
\begin{gather*}
    r_t = r_t^{dense} + r_t^{sparse} \\
    r_t^{dense} = \alpha \times (\frac{1}{d_t + 0.1} - \frac{1}{d_{t-1} + 0.1}) \\
    r_t^{sparse} = \begin{cases}
        \beta, \text{if successfully retrieves the object} \\
        0, \text{otherwise}
    \end{cases}
\end{gather*}
where $\alpha$ and $\beta$ are hyperparameters used to balance the dense reward versus the sparse reward. In our experiments, we set $\alpha = 20$ and $\beta = 800$.

Our policy has one actor network and one critic network, trained with PPO~\cite{schulman2017proximal}. 
Both networks are a 5-layer multilayer perceptron (MLP), and each hidden layer has 256 neurons.
Since our task is not image-based, the actor and critic networks do not share any weights or layers, which is a common practice and shown to have better performance for control tasks~\cite{andrychowicz2020matters}.
We use \texttt{arctan} as our activation function between layers. Notice that we also tried a Transformer~\cite{vaswani2017attention} and an LSTM~\cite{hochreiter1997long} encoder to learn the relationship among observations from different steps in our history buffer; however, we found that the fully connected MLP learns faster and converges to a higher reward.

\subsection{Curriculum Training Strategy}

We train our policy completely in a simulated environment, then attempt zero-shot transfer to real hardware. However, this approach raises the important issue of simulating granular media, which, as noted earlier, is still an active research problem. Current methods for simulating granular media are still slow compared to open-air simulation, sometimes even slower than real-time. However, training inside simulated granular media is still necessary for us due to the gap between the open-air environment and the granular media environment. Apart from extra sensor noise, the object motion differs significantly between the two environments. First, the object can be perturbed by forces propagated through granular media even though the gripper is still some distance away from the object. Second, the object can move in 6DOF in granular media rather than in a 2D plane when sitting on a surface in open air. As a result, policy trained purely in the open-air environment will transfer badly to a granular media environment. 

To not only bridge the gap from open-air to granular media but also train fast, our method uses a curriculum training strategy, as shown in Fig.~\ref{fig:curriculum}. We first train a policy on an open-air tabletop environment and then fine-tune the tabletop policy inside a granular media environment. The tabletop task follows the same RL formulation except that the object sits on a tabletop in the open air. With this curriculum, we are able to bootstrap the training inside granular media by transferring skills learned from the tabletop environments. This curriculum helps our method to converge to a better performance within a shorter training time. 

In order to increase the simulation speed in granular media further, we sacrifice some simulation accuracy for speed. Compared to the pony beads that we use in the real world, we use spheres to approximate their shape. We also make the particles larger in size and then simulate a smaller number of them compared to the real world. Despite the mismatch in the simulated particles, fine-tuning still helps to adapt the tabletop policy to the granular media environments and, consequently, a better sim-to-real transfer. 
\section{Experimental Setup}

\subsection{Real-world Setup} 

\begin{figure}[b]
\centering
\includegraphics[width=0.9\linewidth]{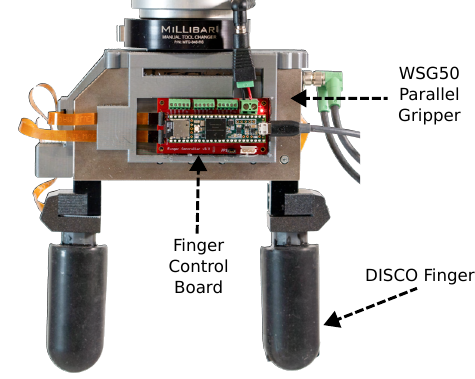}
\caption{\textbf{Parallel gripper with multi-curved tactile fingers.} Our gripper is effective at exploring under granular media due to the DISCO tactile fingers' streamlined shapes and all-around sensing coverage.}
\label{fig:hardware}
\end{figure}

We use a WSG50 parallel gripper with two DISCO~\cite{piacenza2020sensorized} tactile fingers, as shown in Fig.~\ref{fig:hardware}. We mount this gripper to a UR5 robotic arm to control its movement. This setup is effective for interacting with objects under granular materials, as the hemispherical top allows easy penetration into granular media and the cylindrical shape allows smooth movement with small drag. This multi-curved tactile finger has sensing abilities covering the hemisphere top and cylinder. This all-around sensing coverage allows it to sense contact from many directions, which is essential for exploration under granular media. When fully opened, the distance between the two finger centers is 143mm.

\begin{figure*}[h]
\centering
\includegraphics[width=0.95\textwidth]{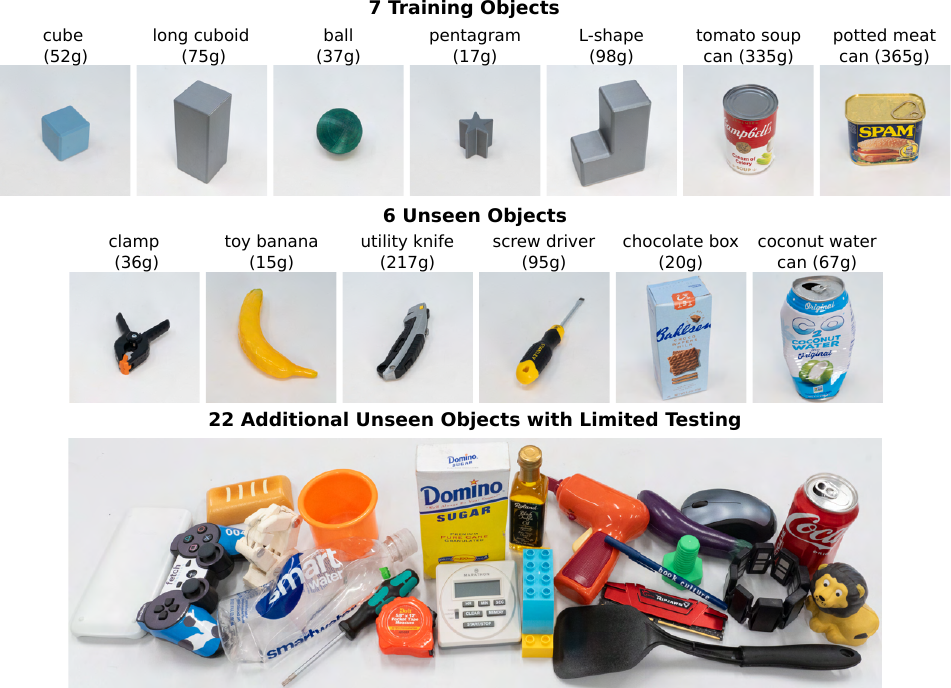}
\caption{\textbf{35 objects used in our real-robot experiments.} Our method is trained on a set of 7 objects. It is then evaluated quantitatively (10 attempts on each object) on both the set of 7 training objects and a set of 6 previously unseen objects. In additional testing, we also show that our method can work on an extended set of 22 unseen objects with at most three attempts per object. The images of training and unseen objects do not correctly reflect their relative sizes. See the demonstration videos for each object on our website for accurate size information.}
\label{fig:objects}
\end{figure*}

The DISCO finger returns a single contact containing the contact location vector and force vector. The tactile finger uses a pre-trained machine learning model to predict a single contact from the high-dimensional raw signals. When there are multiple contacts, the output contact location roughly matches the location of the contact with the largest force, and the output force roughly matches the combined net force from all contact forces.

To filter out ubiquitous contacts with the granular media and keep only the pushing contacts with the target object, we discard those contacts with a force magnitude smaller than 4N. We fill in zeros into the location and force vectors to our policy when there are no pushing contacts detected at a particular step. We find the 4N filtering threshold empirically through calibration, and we use the same 4N threshold for all objects. The optimal filtering threshold should vary across different objects and different types of granular media. But our policy learns to be robust to spurious pushing contacts so we do not need object and granular media-specific calibration. The granular media in our experiments are pony beads with a diameter of 9mm. Our gripper with DISCO fingers is able to move smoothly within such media without jamming. 

In our real-world experiments, we evaluate our method on 35 objects in total, as shown in Fig.~\ref{fig:objects}. Our object set covers a variety of complex shapes, including rigid, deformable, and articulated objects. The object weight in our set also varies a lot, with the lightest object (pen) being 4.5g and the heaviest object (sugar box) being 470g. We evaluate quantitatively on 13 objects with 10 trials each and among them, 7 objects are used for training in simulation and 6 objects are unseen. We then qualitatively show that our method can work on another 22 unseen objects randomly selected from our lab. We quickly go through this set of objects, and we move on to the next object as soon as a successful trial is achieved on the current object. 

The gripper starts initially in the air, and it follows a predefined heuristic trajectory to move into the granular media, makes a shaking motion to shake off jammed beads, and then moves straight forward until the first pushing contact. We start applying force filtering after the finger starts moving forward inside the granular media to filter out ubiquitous contacts, and the learned policy only takes over after the first pushing contact is obtained.

We bury the object so that part of the object occupies the initial forward trajectory of the finger. This is to guarantee an initial first contact as the finger moves forward. Between each trial, we randomize the location of the object, alternate the location of the object in front of the left and right fingers, and we also rotate the object to cover a diverse set of orientations. The episode starts with the first pushing contact and then ends when the maximum steps of 100 are reached or the policy executes a grasp. 


\subsection{Simulation Setup} 
Different from the real-world experiments where we only evaluate inside granular media, we have two different setups for the curriculum training strategy in simulation: tabletop and granular media. We build the two environments using the IsaacGym physics simulator~\cite{makoviychuk2021isaac}. For both environments, we simulate a floating parallel gripper that moves in a 2D plane, according to its real hardware dimensions. 

Despite the fact that we attempt to bury the object such that its center is in front of one of the fingers, it is very hard to guarantee it in practice on the real setup. As a result, in order to be robust to the object location variance, we add more randomization on the object location and orientation during training in simulation. In each episode in the simulation, the object is loaded at a fixed distance away from the gripper. The exact distance does not matter since the finger always moves straight forward until the first contact. However, we vary the location of the object along the direction perpendicular to the gripper's approaching direction by $\pm2$cm. We also randomize the orientation of the object. It is possible that for small objects like pentagrams, the finger can pass the object without making contact under the location randomization, in which case we simply move on to the next episode. 

For simulating granular media, we use spheres with a diameter of 14mm as our particles, and we simulate 4000 particles in total. These spheres do not match the shape of the pony beads in real-world experiments and their sizes are much larger. However, simulating the exact shape and size makes the granular media simulation extremely slow and impractical for training. Despite such a mismatch, fine-tuning in simulated granular media still helps a lot in transferring the tabletop policy to the real robot. 

Unlike the real DISCO finger, which reports a single contact, in simulation we can obtain the contact between the finger and any other object, including the granular beads. We then use a force filter of 3N to discard the ubiquitous contacts. Since there can be multiple pushing contacts even after applying the force filter, we feed the contact with the largest force magnitude into our policy. 

\subsection{Sensor Noise}
\label{sec:noise}
Real-world tactile sensors all show some level of erroneous readings, especially inside granular media. One advantage of the learning-based method is that we can incorporate noise into our training process in simulation for more robust transfer onto the real robot. For a single contact on our DISCO finger, the reported level for the contact location noise is between 1mm and 2mm on average, and the reported level for the contact force noise is 0.1N on average for a force level of $3 \sim 4$N. As a result, due to increased noise level in granular media, we apply a larger location noise uniformly distributed from $[-1\text{cm}, 1\text{cm}]$ and a larger force noise uniformly distributed from $[-0.2\text{N}, 0.2\text{N}]$ to each element in the contact location and net force vectors during training in simulation for all learning-based baselines. 

\subsection{Experimental Conditions and Baselines}

We perform an extensive evaluation of \OURS on the object sets introduced earlier, in both simulation and on real hardware. Since the task we are tackling here has not been previously demonstrated in the literature, we do not have a direct state-of-art method to compare against. However, we compare against a number of alternative approaches and ablations aiming to quantify the importance of each component of our approach. We present all experimental conditions below:

\vspace{3mm}
\noindent \textbf{Align-to-grasp (A2G)}.
This method rotates the gripper to align with the direction of the contact force after the first pushing contact is detected. This is a heuristics-based method and does not require training, intended to evaluate the importance of a learning-based method that trains in the presence of noise. 

\vspace{3mm}
\noindent \textbf{A2G-push}.
This variant of A2G starts rotating and aligning only after five consecutive contacts on the same finger have been detected. This is intended to filter out spurious pushing contacts in granular media, and to evaluate the performance of pushing alone as a heuristic for reducing uncertainty.

\vspace{3mm}
\noindent \textbf{\OURS-tabletop}.
This is the policy from the first stage of our curriculum training pipeline, trained on the open-air tabletop environment. 

\vspace{3mm}
\noindent \textbf{\OURS-granular}.
This is the policy defined in our method, but trained exclusively in granular media, thus bypassing our proposed curriculum.

\vspace{3mm}
\noindent \textbf{\OURS}.
This is our full proposed method, combining all the aspects discussed in this study.

\section{Results and Discussion}

\begin{figure*}[h!]
\centering
\includegraphics[width=\linewidth]{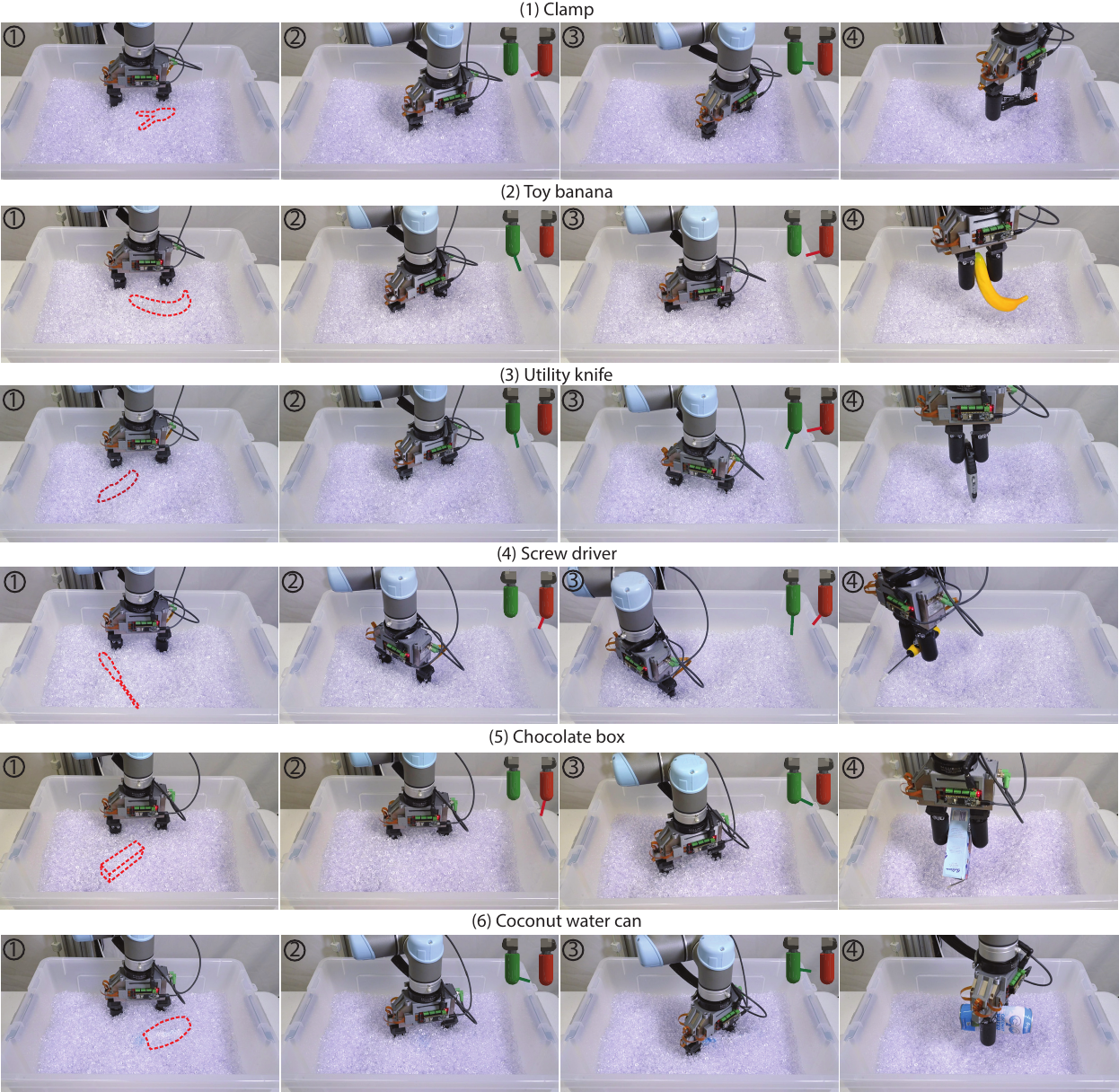}
\caption{\textbf{\OURS successful demonstrations on the real robot.} We show one example each for the six unseen objects evaluated quantitatively. The first image in each example outlines the initial position of the buried object. In the second and third images, our policy interacts with the object using the emergent pushing actions. We visualize the contact locations and normals on the rendered fingers. Our policy moves the object to a graspable location, and in the fourth image, the objects are successfully retrieved. 
Video demonstrations can be found on our project website at \texttt{\url{https://jxu.ai/geotact}}.}
\label{fig:real_demo}
\end{figure*}

In this section, we present and discuss the results of our real-robot and simulation experiments. Visit our project website at \texttt{\url{https://jxu.ai/geotact}} for video demonstrations and additional information. 

\subsection{Real-robot Results}

\begin{table*}
    \caption{\textbf{Real-world experiment results on 7 training objects.} We show the number of successes out of 10 trials.}
    \centering
    \begin{tabular}{c|ccccccc|c}
    \toprule
    \multirow[b]{2}{*}{\textbf{Method}} & \multicolumn{7}{c}{\textbf{Training Objects}} & \multirow[b]{2}{*}{\textbf{Average}}\\
    & Cube & \makecell{Long\\Cuboid} & Ball & Pentagram & L-shape & \makecell{Tomato\\Soup Can} & \makecell{Potted\\Meat Can} \\
    \midrule
    A2G & 3/10 & 3/10 & 0/10 & 2/10 & 4/10 & 1/10 & 4/10 & 0.24 \\
    \OURS & \textbf{7/10} & \textbf{8/10} & \textbf{1/10} & \textbf{6/10} & \textbf{6/10} & \textbf{3/10} & \textbf{7/10} & \textbf{0.54} \\
    \bottomrule
    \end{tabular}
    \label{tab:real-train}
\end{table*}

\begin{table*}
    \caption{\textbf{Real-world experiment results on 6 unseen objects.} We show the number of successes out of 10 trials.}
    \centering
    \begin{tabular}{c|cccccc|c}
    \toprule
    \multirow[b]{2}{*}{\textbf{Method}} & \multicolumn{6}{c}{\textbf{Unseen Objects}} & \multirow[b]{2}{*}{\textbf{Average}} \\
    & Clamp & \makecell{Toy\\Banana} & \makecell{Utility\\Knife} & \makecell{Screw\\Driver} & \makecell{Chocolate\\Box} & \makecell{Coconut\\Water Can} \\
    \midrule
    A2G & 3/10 & 3/10 & 3/10 & 2/10 & 4/10 & 4/10 & 0.32 \\
    \OURS & \textbf{8/10} & \textbf{6/10} & \textbf{6/10} & \textbf{5/10} & \textbf{7/10} & \textbf{6/10} & \textbf{0.63} \\
    \bottomrule
    \end{tabular}
    \label{tab:real-unseen}
\end{table*}

We evaluate two representative methods on the real robot --- our method, \OURS, which is learning-based and uses pushing, and A2G, which is heuristics-based and does not use pushing. We first evaluate both methods on the training set (seven objects) and unseen set (six objects) shown in Fig.~\ref{fig:objects}, performing ten trials on each object. The results on the training set and the unseen set are shown in Table~\ref{tab:real-train} and Table~\ref{tab:real-unseen}, respectively. 

\OURS achieves success rates of 54\%  and 63\% on the training set and unseen set, respectively, demonstrating strong generalization to unseen objects without any performance drop.
It outperforms A2G, which suffers from the significant amount of sensor noise introduced by granular media. In comparison, \OURS is more robust to sensing noise by continuously interacting with the object through emergent pushing actions and reacting to new contact observations. In addition, fine-tuned in granular media, \OURS is more adapted to the object motion discrepancy between tabletop and granular media. We also notice that the success rates of \OURS on two predominantly rounded shapes (ball and tomato soup can) are significantly lower compared to other objects. We discuss these failure cases, along with failure cases of A2G, later in Sec.~\ref{sec:failure}.

An example of \OURS's noise-robustness behavior can be seen in the chocolate box demonstration in Fig.~\ref{fig:real_demo}. There is a spurious pushing contact reading from the right (red) tactile finger even though the object is still some distance away. As our policy reacts to that spurious reading, the left (green) finger obtains a true contact. Our policy then pushes the object for a successful grasp. 

In additional testing on an extensive set of objects, we test our method on the set of 22 novel objects (not used in training) shown in Fig.~\ref{fig:objects}. For each object, we try \OURS until we obtain a successful retrieval. All 22 objects are successfully retrieved within at most three attempts. 

\subsection{Simulation Results}

\begin{table*}
    \addtolength{\tabcolsep}{-2pt}
    \renewcommand\cellset{\renewcommand\arraystretch{0.6}}
    \caption{\textbf{Simulation experiment results on the tabletop.} Each success rate is computed over 1000 trials.}
    \centering
    \begin{tabular}{c|ccccccc|c}
    \toprule
    \multirow[b]{2}{*}{\textbf{Method}} & \multicolumn{7}{c|}{\textbf{Training Objects}} & \multirow[b]{2}{*}{\textbf{Average}} \\
    & Cube & \makecell{Long\\Cuboid} & Ball & Pentagram & L-shape & \makecell{Tomato\\Soup Can} & \makecell{Potted\\Meat Can} \\
    \midrule
    A2G & 0.61 & 0.53 & 0.86 & 0.46 & 0.45 & 0.90 & \textbf{0.77} & 0.65 \\
    \OURS-tabletop & \textbf{0.96} & \textbf{0.80} & \textbf{1.00} & \textbf{0.96} & \textbf{0.89} & \textbf{0.97} & \textbf{0.77} & \textbf{0.91} \\
    \bottomrule
    \end{tabular}
    \label{tab:sim-tabletop}
\end{table*}

\begin{table*}
    \addtolength{\tabcolsep}{-2pt}
    \renewcommand\cellset{\renewcommand\arraystretch{0.6}}
    \caption{\textbf{Simulation experiment results inside granular materials.} Each success rate is computed over 1000 trials.}
    \centering
    \begin{tabular}{c|ccccccc|c}
    \toprule
    \multirow[b]{2}{*}{\textbf{Method}} & \multicolumn{7}{c|}{\textbf{Training Objects}} & \multirow[b]{2}{*}{\textbf{Average}}\\
    & Cube & \makecell{Long\\Cuboid} & Ball & Pentagram & L-shape & \makecell{Tomato\\Soup Can} & \makecell{Potted\\Meat Can} \\
    \midrule
    A2G & 0.22 & 0.54 & 0.49 & 0.21 & 0.45 & 0.48 & 0.35 & 0.39 \\
    A2G-push & 0.50 & 0.60 & 0.48 & 0.19 & 0.51 & 0.44 & 0.36 & 0.44 \\
    \OURS-tabletop & 0.18 & 0.41 & 0.13 & 0.08 & 0.73 & 0.06 & 0.33 & 0.27 \\
    \OURS-granular & 0.73 & 0.74 & 0.67 & 0.56 & \textbf{0.77} & 0.43 & 0.31 & 0.60 \\
    \OURS & \textbf{0.79} & \textbf{0.84} & \textbf{0.73} & \textbf{0.60} & 0.71 & \textbf{0.53} & \textbf{0.50} & \textbf{0.67} \\
    \bottomrule
    \end{tabular}
    \label{tab:sim-granular}
\end{table*}

We present more extensive experiments and ablations to analyze our method and the baselines in simulation, which lends itself to experiments and analysis on a larger scale.

First, in an open-air tabletop environment, we evaluate A2G and \OURS-tabletop using the same amount of noise as used in training (shown in Table~\ref{tab:sim-tabletop}). Note that we do not evaluate A2G-push on the tabletop, as A2G-push is designed to be robust to spurious contact readings, which do not happen in an open-air environment. Without the extra sensor noise and complex object motion introduced by granular media, \OURS-tabletop is achieving over 90\% success rate. A2G also has a higher success rate (65\%) compared to its evaluation in granular media on the real robot. However, A2G struggles with a few objects, such as L-shape, long cuboid, and pentagram, even in the tabletop environment, and we discuss these failure cases in detail in the next section. 

Inside granular media simulation, we evaluate all five methods using the same amount of tactile noise as training (Table~\ref{tab:sim-granular}). Working within granular media is much more challenging than the tabletop setup. As a result, the performance of both \OURS-tabletop and A2G drops significantly compared to tabletop simulation, from 91\% to 27\% and from 65\% to 39\%, respectively. A2G-push performs slightly better (44\%) than A2G, showing that adding heuristic push can reduce uncertainty and increase robustness.

The importance of fine-tuning inside simulated granular media can be shown by the performance improvement from \OURS-tabletop (27\%) to \OURS (67\%). Fine-tuning helps our policy adapt to the larger sensor noise and object motion discrepancy in granular media.

\begin{figure}[h]
\centering
\includegraphics[width=0.95\linewidth]{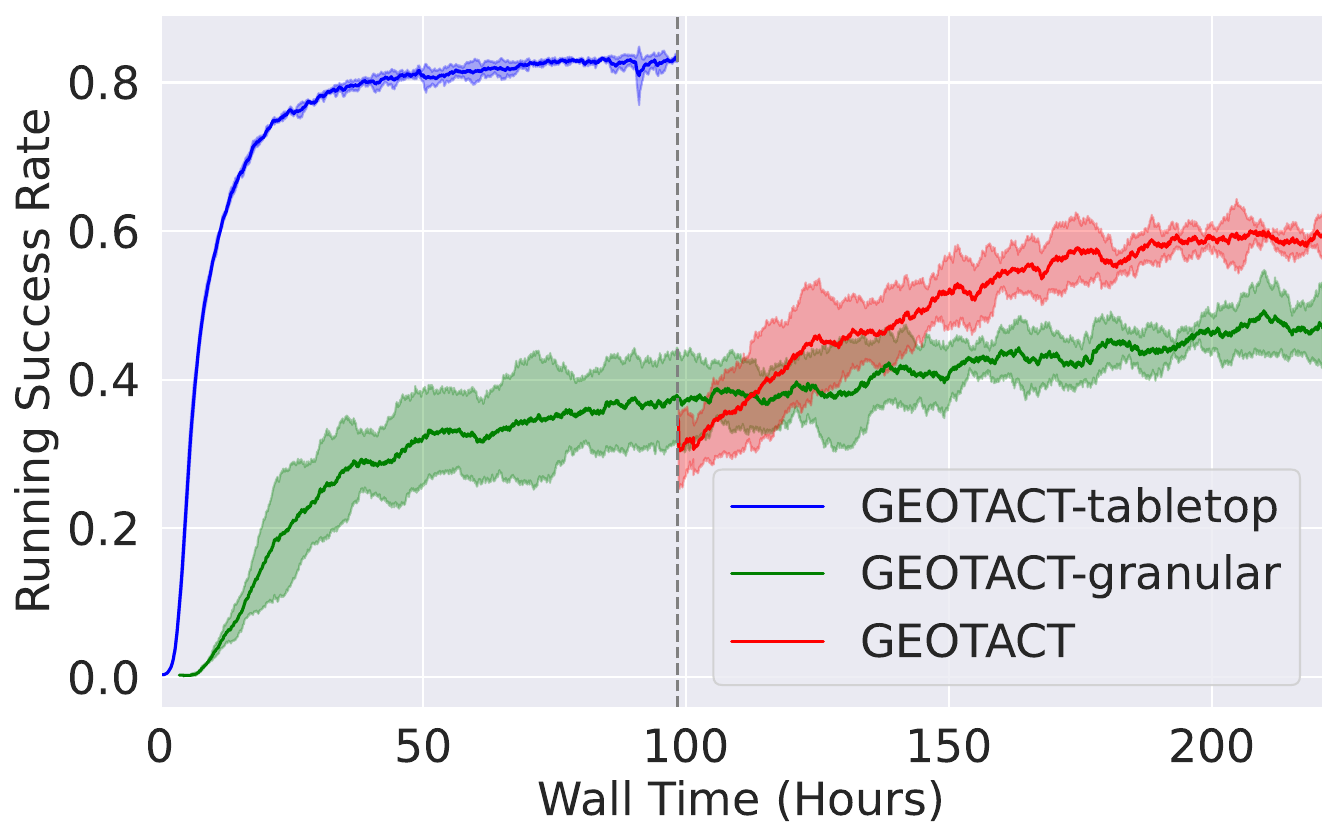}
\caption{\textbf{Training plots of the curriculum strategy vs no-curriculum baseline.} We show the running success rate (mean $\pm$ standard deviation) of the past 1,000 episodes during training, generated over five random seeds. Despite the initial performance drop when \OURS-tabletop is loaded into the granular media environment (indicated by the dashed line), \OURS quickly surpasses \OURS-granular and converges to a better performance.}
\label{fig:curriculum_training_plot}
\end{figure}

\begin{figure}[h]
\centering
\includegraphics[width=0.95\linewidth]{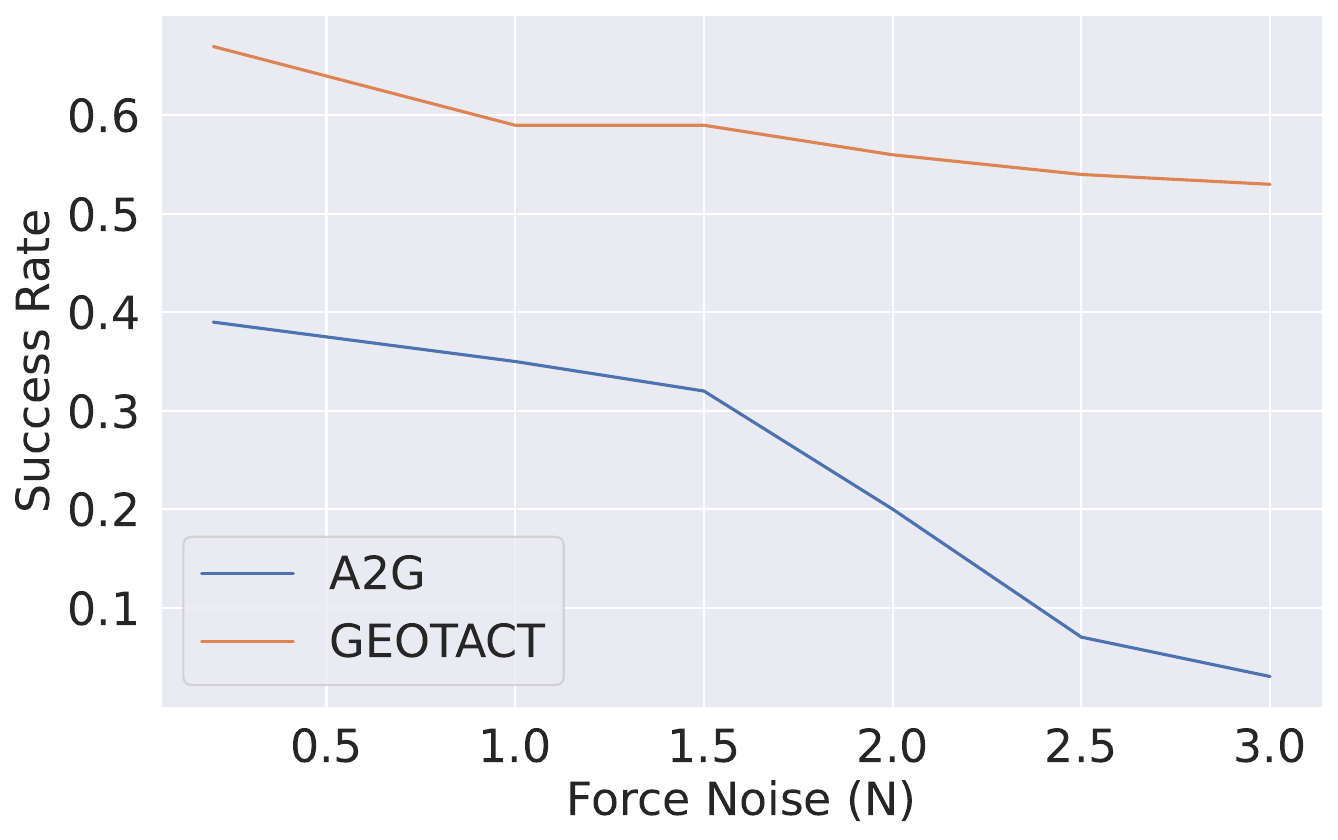}
\caption{\textbf{Noise robustness analysis in granular media simulation.} When evaluated on force noise larger than training, \OURS maintains above 50\% success rate even with 3N force noise, while A2G's performance drops to 3\%.}
\label{fig:noise}
\end{figure}

The performance gap between \OURS (67\%) and \OURS-granular (60\%) demonstrates the necessity of the proposed curriculum training strategy. Without pre-training on the tabletop, \OURS-granular performs worse than \OURS. We further show in Fig.~\ref{fig:curriculum_training_plot} the training plots of \OURS and \OURS-granular. After the initial performance drop when the tabletop policy is loaded into the granular media, the success rate of \OURS quickly surpasses \OURS-granular and eventually converges to higher performance. This experiment shows that pre-training on the tabletop is able to boost the training speed and improve the final performance of our policy. 

Finally, we then evaluate the noise robustness of \OURS, and compare it against A2G when faced with sensor noises larger than seen in training. 
We gradually increase the force noise range up to [-3N, 3N], and the success rate is shown in Fig.~\ref{fig:noise}. At a force noise level of [-3N, 3N], the success rate of A2G drops to 3\% while \OURS remains over 50\%. As an open-loop policy, A2G relies heavily on an accurate first contact. However, when the previous contact information is largely incorrect, \OURS is reactive and can continue interacting with the target object through emergent pushing behaviors.

\subsection{Failure Cases and Discussion}
\label{sec:failure}

\begin{figure*}[h]
\centering
\includegraphics[width=\linewidth]{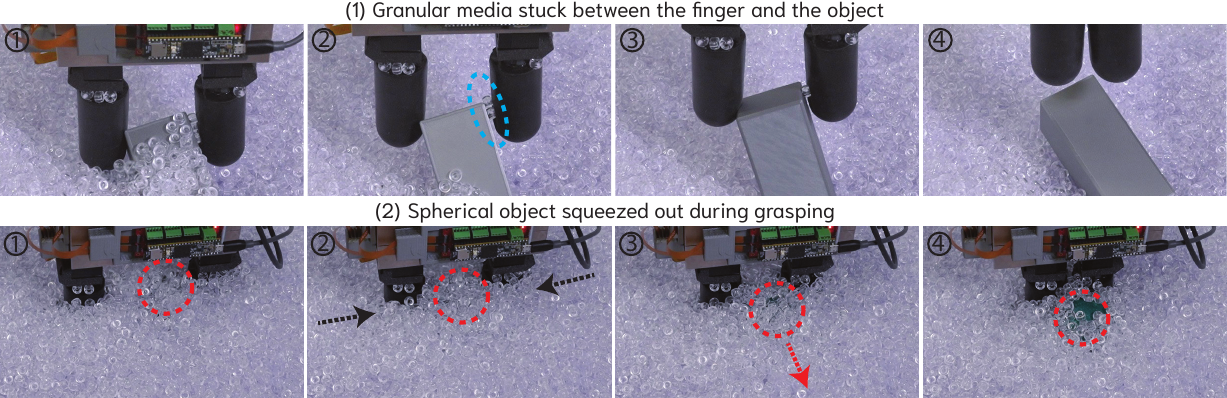}
\vspace{-3mm}
\caption{\textbf{\OURS failure case demonstrations on the real robot.} (1) There is a thin layer of granular beads (circled in blue) stuck between the finger and the object during grasping. Both the plastic surface and the beads are slippery, so the grasp fails. (2) Spherical and cylindrical objects are very hard to grasp with our parallel gripper with cylindrical fingers. They are always squeezed out even though a good grasp is formed. The location of the sphere is marked in red circles. The movement of the fingers and the sphere is marked with dashed lines with arrowheads.}
\label{fig:failure_cases}
\end{figure*}

\begin{figure}[!h]
\centering
\includegraphics[width=0.98\linewidth]{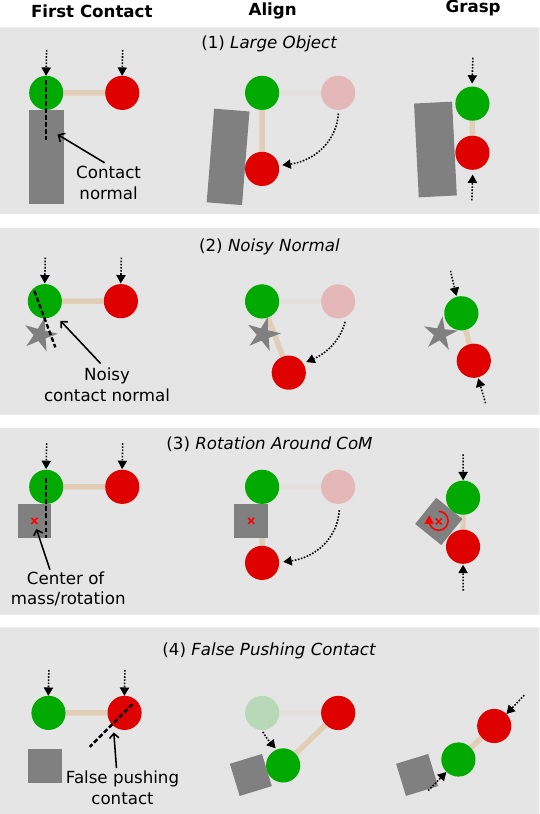}
\vspace{2mm}
\caption{\textbf{Bird's-eye view of four failure scenarios for A2G.} A2G consists of three stages: obtaining the first contact, aligning the gripper with the contact normal, and then closing the gripper. See Sec.~\ref{sec:failure} for a detailed explanation of these failure cases. 
}
\label{fig:a2g_failure}
\end{figure}

We notice that our method often fails on rounded shapes (e.g., ball, tomato soup can) when deployed on the real robot, despite a high success rate on the same objects in simulation. A successful grasp of these two objects is extremely sensitive to sensor error, due to their rounded shape. On the real robot, our policy locates both objects well, but the gripper simply often squeezes them out during closing, as shown in Fig.~\ref{fig:failure_cases}(2). A very precise grasp (centered on the object) is necessary, which is easier to achieve in a simulated environment with perfect control and low sensor noise. This is also partly an inherent limitation of a parallel gripper with cylindrical fingers attempting to grasp a round object. A multi-fingered hand might work better for such objects, and we leave it for future work.

The two objects, ball and tomato soup can, is exactly the reason why our method has a higher success rate on unseen objects than training objects. Despite the fact that these two objects are seen during training, they are inherently hard to grasp due to the limitation of the parallel gripper with cylindrical fingers. If we exclude these two objects, the remaining objects in the training set actually have a higher average success rate than unseen objects. Most of our unseen objects have irregular shapes and are not cylindrical or spherical, which is not challenging from a hardware perspective. The coconut water can was originally cylindrical, but it was squeezed and lost its perfect cylindrical shape.

Another common failure case of our policy on the real robot is due to a thin layer of granules stuck between the target object and the finger, as shown in Fig.~\ref{fig:failure_cases}(1). When this happens, due to the low friction coefficient with the granules, the object will slip out of the formed grasp. Again, this is partly a limitation of our parallel gripper, and a multi-fingered gripper might be able to relieve this problem. We will leave this to future work. 

A2G also fails for the same two reasons above, but in addition, it fails in another four cases, as shown in Fig.~\ref{fig:a2g_failure}. (1) \textit{Large Object}: For large objects such as L-shape or long cuboid, because the maximum length of the object exceeds the span of the gripper, the gripper can only grasp successfully along the short dimension. Simply aligning the gripper with the contact normal is likely to push the object away. (2) \textit{Noisy Normal}: The contact normal can be noisy, especially when contacting a pointy/sharp edge of an object, such as that of the pentagram, making the aligned grasp unstable. This failure case happens much more often in granular media than on the tabletop. (3) \textit{Rotation Around CoM}: During the grasping phase of A2G, the finger contacting the object will start pushing the object, and then the object will start rotating around its center of mass. This rotation can result in the object sliding outside the grasping closure. (4) \textit{False Pushing Contact}: When there is a spurious pushing contact being reported by the finger, A2G does not have any recovery behavior and will align to the force direction of the false contact, leading to an improper grasp. This failure case is specific to granular media since there is no ubiquitous contact to filter on the tabletop.

In comparison, \OURS is resilient to all these failure types. We attribute this to the combination of its reactive nature, obtained via end-to-end training in the presence of noise, and the emergent pushing actions that reduce the uncertainty of the object location and funnel the object into a stable grasp.

\section{Limitations and Future Works}

\addedtext{1-2}{We identify some limitations and important future directions of this work, listed below.

\textbf{Hardware improvement.} Common failure cases, such as spherical and ball-shaped objects, are extremely challenging with a parallel gripper, and moving to a multi-fingered hand is an important future direction. Even for humans, retrieving buried objects with only two fingers is a challenging task, putting a realistic upper bound on the performance with parallel grippers. In addition, with a multi-fingered hand, we can explore more complicated action spaces and action primitives, such as scooping.

\textbf{More complicated action space.} 
Our method restricts the gripper movement inside the granular media to a 2D plane, and future work includes exploring more complicated action spaces (such as 6DOF grasp at an angle) and diverse action primitives (such as scooping, especially with multi-fingered hands). The expanded action space will also require a more efficient training pipeline.

\textbf{More challenging tasks.}
We made some assumptions to simplify the task, such as a guaranteed first touch and single object retrieval. In future directions, we want to relieve the assumptions by incorporating initial searching and finding policies to actively locate the target object. We also want to study the retrieval of a particular object among many buried objects. These are more challenging and similar to real-world scenarios, and will likely require a combination of RL training and additional heuristic guidance.

\textbf{Generalization to other granular types.} 
Our experiments involve one type of granular media, and in our future work, we aim to evaluate our method on other types of granules. We note, however, that our beads have relatively large, non-spherical shapes and display anisotropic behavior. Sands and soil, which have much smaller granules and more isotropic properties, might make it easier to distinguish pushing contact from ubiquitous contact. Other failure modes, such as objects sliding out of a grasp due to a layer of granules stuck between fingers and objects, might also be less frequent. 
\modifiedtext{1-4}{It is also worth noting that smaller granules have less compressibility, potentially generating higher resistance forces for both ubiquitous contact and pushing contact. This necessitates the recalibration of the force threshold to filter ubiquitous contacts.}

We also speculate that larger irregular gravel or heterogeneous granular media will be more challenging than the granular beads that we tested on. There will be more jamming for larger or irregular gravels, and the pushing contact will be more difficult to distinguish from ubiquitous contact, meaning a larger gap between open-air policy and granular media policy. In fact, if we keep increasing the size of the granules, such that the granules and the object are of similar size, then the task will become more of a targeted object singulation task.
Of course, these hypotheses will need to be rigorously tested in future work. 


}

\section{Conclusion}

In this work, we propose \OURS, a novel method that reliably retrieves objects buried completely under granular media using only touch sensing. This is a challenging task due to the absolute absence of vision and spurious and noisy tactile readings, introduced by the ubiquitous contacts inside granular media. We formulate this task as a model-free reinforcement learning problem and train it end-to-end in simulation with a curriculum strategy. We first train the policy in an open-air environment and then fine-tune it inside a granular environment. Our formulation of the learning task action space leads to emergent pushing behaviors that help reduce uncertainty and also move the object to form a stable grasp. Compared to heuristic-based methods, our learned policy can incorporate simulated sensor noise during training and is thus more robust to spurious and noisy tactile readings. Despite being trained only on 7 objects with simple shapes in simulation, we show that our method can zero-shot transfer to the real hardware and successfully pick up 28 unseen objects, including rigid, deformable, and articulated objects with various complex shapes. 

\backmatter

\bmhead{Acknowledgements}
We would like to thank Eric Chang and Amr El-Azizi for their support on the DISCO tactile finger. We would like to thank Zhenjia Xu for his help with the granular media simulation. This material is based upon work supported by the National Science Foundation under Award No. CMMI-2037101, and the Office of Naval Research under grant N00014-21-1-4010. Any opinions, findings and conclusions or recommendations expressed in this material are those of the author(s) and do not necessarily reflect the views of the National Science Foundation or the Office of Naval Research.




\bibliography{references}

\end{document}